%% file: paper.tex
\documentclass[runningheads]{llncs}
\usepackage{graphicx}
\usepackage{amsmath}
\usepackage{amssymb}
\usepackage{booktabs}
\usepackage{threeparttable}
\usepackage{subfig}
\usepackage{pifont}
\usepackage[pagebackref,breaklinks,colorlinks]{hyperref}
\usepackage{svg}
\usepackage[capitalize]{cleveref}
\crefname{section}{Sec.}{Secs.}
\Crefname{section}{Section}{Sections}
\Crefname{table}{Table}{Tables}
\crefname{table}{Tab.}{Tabs.}

\begin{document}

\def\ie{\emph{i.e}\onedot}
\newcommand{\zhccr}{Zero-shot HCCR }
\newcommand{\DS}{Diffusion Synthesis }
\newcommand{\ds}{diffusion synthesis }
\newcommand{\x}{\mathbf{x}}
\newcommand{\cc}{\mathbf{c}}
\newcommand{\y}{\mathbf{y}}
\newcommand{\g}{\mathbf{g}}
\newcommand{\w}{\mathbf{w}}
\title{Zero-shot Generation of Training Data with Denoising Diffusion Probabilistic Model for Handwritten Chinese Character Recognition}
%
\titlerunning{Zero-shot Generation of Training Data with DDPM for HCCR}
%
%
\authorrunning{Dongnan Gui, Kai Chen, Haisong Ding and Qiang Huo}
%
\author{Dongnan Gui\inst{1,2}$^{,}$\thanks{This work was done when Dongnan Gui was an intern in MMI Group, Microsoft
Research Asia, Beijing, China.} \and
Kai Chen\inst{1}$^{,}$\thanks{Corresponding author.} \and {Haisong Ding}\inst{1}$^{}$ \and Qiang Huo\inst{1}}
\institute{Microsoft Research Asia, Beijing, China \and
University of Science and Technology of China, Hefei, China 
\\
\email{gdn2001@mail.ustc.edu.cn \quad chenkai.cn@hotmail.com \quad dinghs11@mail.ustc.edu.cn \quad qianghuo@microsoft.com}}

\maketitle              
%

%
%
%

\input{0abstarct.tex}
\input{1intro.tex}
\input{2related.tex}
\input{3method.tex}
\input{4experiments.tex}
\input{5future.tex}
\input{6conclusion.tex}

\newpage
\bibliographystyle{splncs04}
\bibliography{egbib}

\end{document}

%% file: 0abstarct.tex
\begin{abstract}
   There are more than 80,000 character categories in Chinese while most of them are rarely used. To build a high performance handwritten Chinese character recognition (HCCR) system supporting the full character set with a traditional approach, many training samples need be collected for each character category, which is both time-consuming and expensive. In this paper, we propose a novel approach to transforming Chinese character glyph images generated from font libraries to handwritten ones with a denoising diffusion probabilistic model (DDPM). Training from handwritten samples of a small character set, the DDPM is capable of mapping printed strokes to handwritten ones, which makes it possible to generate photo-realistic and diverse style handwritten samples of unseen character categories.  Combining DDPM-synthesized samples of unseen categories with real samples of other categories, we can build an HCCR system to support the full character set. Experimental results on CASIA-HWDB dataset with 3,755 character categories show that the HCCR systems trained with synthetic samples perform similarly with the one trained with real samples in terms of recognition accuracy. 
   The proposed method has the potential to address HCCR with a larger vocabulary.
    \keywords{Denoising Diffusion Probabilistic Model \and Handwritten Chinese Character Recognition \and Zero-shot Generation.}
\end{abstract}

%% file: 1intro.tex
\section{Introduction}
\label{sec:intro}

In the latest National Standards of the People's Republic of China about Chinese coded character set (GB18030-2022), 87,887 Chinese character categories are included.
To create a high-performance handwritten Chinese character recognition (HCCR) system that supports the full character set using traditional approaches, a large number of training samples with various writing styles would be collected for each character category.
However, only about 4,000 categories are commonly used in daily life. It is therefore both time-consuming and expensive to collect representative handwritten samples for the remaining 95\% rarely-used ones. 
These categories are often of complicated structures, existing in personal names, addresses, ancient books, historic documents and scientific publications. 
An HCCR system supporting the full-set of these categories with high accuracy will be beneficial to improve user experience, protect cultural heritages and promote academic exchanges.

Lots of research efforts have been made to build an HCCR system with only real training samples from commonly used characters. A Chinese character consists of radicals/strokes with specific spatial relationships, which are shared across all characters. Rather than encoding each character category as a single one-hot vector, \cite{wang2018denseran,cao2020zero,wang2019radical,diao2022rezcr} encode it as a sequence of radicals/strokes and spatial relationships to achieve zero-shot recognition goal. In \cite{ao2019cross,liu2022open,liu2022towards,huang2021zero}, font-rendered glyph images are leveraged to provide reference representations for unseen character categories.  There are also some efforts to synthesize handwritten samples for unseen categories. For example, \cite{xue2021radical} synthesizes unseen character samples with a radical composition network and combines them with real samples to train an HCCR system. However, its recognition accuracy is relatively poor.

We propose to solve this problem by synthesizing diverse and high-quality training samples for unseen character categories with denoising diffusion probabilistic models (DDPMs) \cite{sohl2015deep,ho2020denoising}.
Diffusion models have been shown to outperform other generation techniques in terms of diversity and quality~\cite{nichol2021improved,dhariwal2021diffusion,song2019generative,song2020improved,song2020score}, due to their powerful modeling capacity of high-dimensional distributions. This also offers a zero-shot generation capability. 
For example, in diffusion-based text-to-image generation~\cite{nichol2021glide,ramesh2022hierarchical,saharia2022photorealistic}, with all object types and spatial relationships existed in training samples, diffusion models are capable of generating photo-realistic images of in-existence object combinations and layouts.
As mentioned above, Chinese characters can be treated as combinations of different radicals/strokes with specific layouts. We can leverage DDPM to achieve the goal of zero-shot handwritten Chinese character image generation.

In this paper, we design a glyph conditional DDPM (GC-DDPM), which concatenates a font-rendered character glyph image with the original input of U-Net used in ~\cite{dhariwal2021diffusion}, to guide the model in constructing mappings between font-rendered and handwritten strokes/radicals. To the best of our knowledge, we are the first to apply DDPMs to zero-shot handwritten Chinese character generation.
Unlike other image-to-image diffusion model frameworks~(e.g., \cite{saharia2022palette,wang2022pretraining,pang2021image}), which aim at synthesizing images in the target domain while faithfully preserving the content representations, our goal is to learn mappings from rendered printed radicals/strokes to the handwritten ones. 

%
%
Experimental results on CASIA-HWDB~\cite{liu2011casia} dataset with 3,755 character categories show that the HCCR systems trained with DDPM-synthesized samples outperform other synthetic data based solutions and perform similarly with the one trained with real samples in terms of recognition accuracy.
We also visualize the generation effect of both in and out of 3,755 character categories, which indicates that our method has the potential to be extended to a larger vocabulary.

The remainder of the paper is organized as follows. In \cref{sec:related work}, we briefly review related works. In \cref{sec:approach}, we describe our GC-DDPM design along with sampling methods. Our approach is evaluated and compared with prior arts in \cref{sec:experiments}. We discuss limitations of our approach and future work in \cref{sec:future}, and conclude the paper in \cref{sec:conclusion}.

%% file: 2related.tex
\section{Related Work}
\label{sec:related work}

\textbf{Zero-shot HCCR}
    Conventional HCCR systems~\cite{cirecsan2015multi,yin2013icdar,zhong2015high,chen2015beyond,zhong2016handwritten,li2018building}, although achieving superior recognition accuracy, can only recognize character categories that are observed in the training set. Zero-shot HCCR aims to recognize handwritten characters that are never observed.
    Most of the previous zero-shot HCCR systems can be divided into two categories: structure-based and structure-free methods. 
    In structure-based methods, a Chinese character is represented as a  sequence of composing radicals~\cite{wang2018denseran,cao2020zero,wang2019radical,diao2022rezcr} or strokes~\cite{chen2021zero}. Although the character is never observed, the composing radicals, strokes and their spatial relationships have been observed in the training set. Therefore, structure-based methods are able to predict the radical or stroke sequences of unseen Chinese characters and achieve zero-shot recognition. 
    However, in these methods, the radical or stroke sequence representations of Chinese characters require lots of language-specific domain knowledge.
    In structure-free method, ~\cite{ao2019cross,liu2022open,liu2022towards,huang2022hippocampus} leverage information from the corresponding Chinese character glyph images. Zero-shot HCCR is achieved by choosing the Chinese character whose glyph features are closest to that of the handwritten ones in terms of visual representations. In \cite{huang2021zero}, the radical information is also used to extract the visual representations of glyph images.

    \textbf{Zero-shot Data Synthesis for HCCR}
    Besides designing zero-shot recognition systems, there are some studies to directly synthesize handwritten training samples for unseen categories. 
    ~\cite{xue2021radical} investigates a radical composition network to generate unseen Chinese characters by integrating radicals and their spatial relationships.
    Although the generated handwritten Chinese characters can increase the recognition rate of unseen handwritten characters, the overall recognition performance is relatively poor.
    In this work, we propose to use a more powerful diffusion model to generate unseen handwritten Chinese characters given corresponding glyph images.

\textbf{Zero-shot Chinese Font Generation} 
    Zero-shot Chinese font generation aims to generate font glyph for unseen Chinese characters based on some seen character / font glyph pairs. 
    In ~\cite{zhang2018separating,gao2019artistic,zhu2020few,xie2021dg,liu2022fonttransformer}, the image-to-image translation framework is used to achieve this goal. 
    Works in ~\cite{huang2020rd,park2021few,liu2022xmp} also leverage the information of composing components, radicals, strokes for better generalization.
    In this paper, we focus on zero-shot handwritten Chinese character generation with DDPM and we can easily adapt this method to zero-shot Chinese font generation task.

\textbf{Diffusion Model} 
    DDPM~\cite{sohl2015deep,ho2020denoising} has become extremely popular in computer vision and achieves superior performance in image generation tasks. DDPM uses two parameterized Markov chains and variational inference method to reconstruct the data distribution.
    DDPMs have demonstrated their powerful capabilities to generate high-quality and high-diversity images ~\cite{ho2020denoising,dhariwal2021diffusion,song2020score}. It is shown in ~\cite{ramesh2022hierarchical} that DDPM can perform a great effect on combination of concepts, which can integrate multiple elements. Diffusion models are also applied to other tasks~\cite{yang2022diffusion,croitoru2022diffusion}, including high-resolution generation~\cite{rombach2022high}, image inpainting ~\cite{wang2022pretraining}, natural language processing~\cite{austin2021structured} and so on. Besides, ~\cite{luhman2020diffusion} introduces DDPM to solve the problem of online English handwriting generation. In this work, we propose to leverage DDPM for zero-shot handwritten Chinese character generation and to synthesize training data for unseen Chinese characters to build HCCR systems.

%% file: 3method.tex
\section{Our Approach}
\label{sec:approach}
\subsection{Preliminary}
\begin{figure*}[!t]
		\centering
		\includegraphics[width=\linewidth]{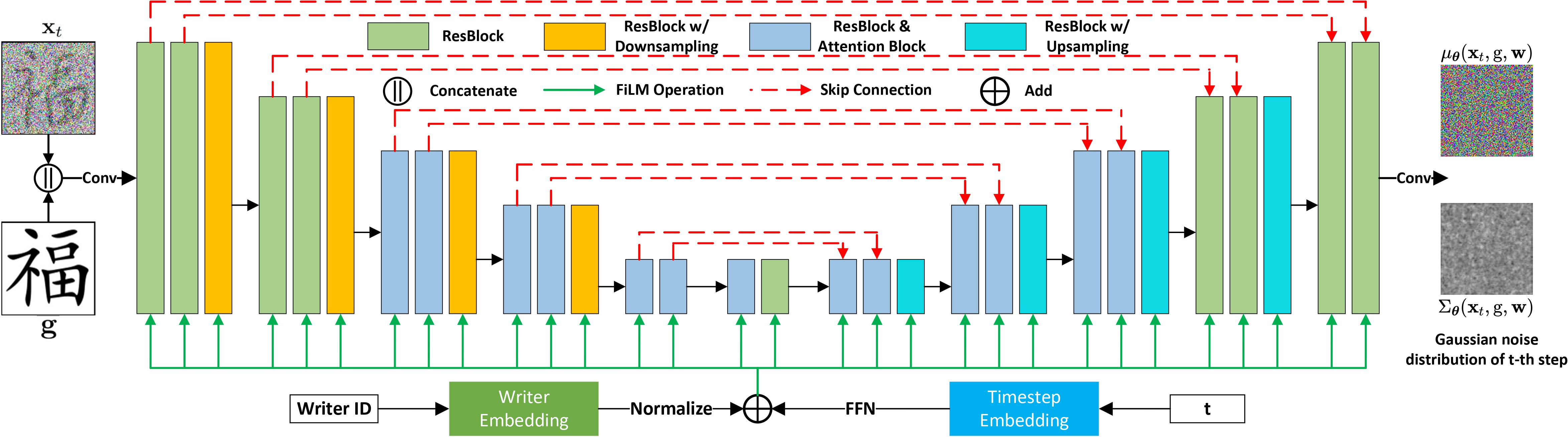}
		\caption{Architecture of glyph conditional U-Net, which is adapted from the model used in \cite{dhariwal2021diffusion}. We concatenate font ``kai'' rendered character image with original input to provide glyph guidance during generation.}
		\label{fig:unet}
        \vspace{-2mm}
\end{figure*}

\begin{figure}[thbp]
		\centering
        \vspace{-2mm}
		\includegraphics[width=0.8\linewidth]{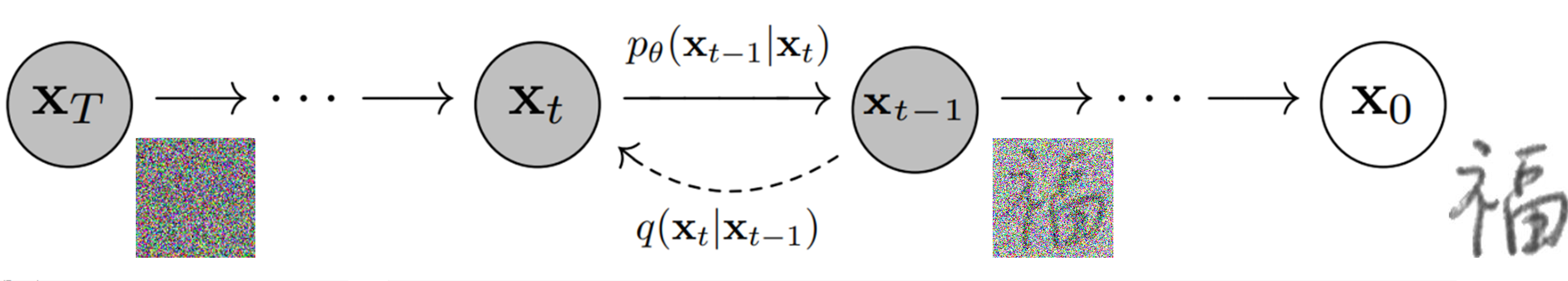}
		\caption{The Markov chain of forward (reverse) diffusion process of generating a handwritten Chinese character sample by slowly adding (removing) noise. Adapted from \cite{ho2020denoising}.}
		\label{fig:ddpm}
        \vspace{-2mm}
\end{figure}

Diffusion model is a new paradigm of data generation. It defines a Markov chain of diffusion steps to slowly add random noise to data and then learn to reverse the diffusion process to construct desired data samples from the noise \cite{weng2021diffusion}. 
As shown in \cref{fig:ddpm}, in our handwritten Chinese character generation scenario, we first sample a character image from the real distribution $\mathbf{x}_0\sim q(\mathbf{x})$.
Then, in forward diffusion process, small amounts of Gaussian noise are added to the sample in steps according to \cref{eqn:diff_forward},
\begin{align}
\label{eqn:diff_forward}
   & q(\x_{t} | \x_{t-1}) =\mathcal{N}(\x_{t} ; \sqrt{1-\beta_{t}} \x_{t-1}, \beta_{t} \mathbf{I})\\
   & \x_{t}= \sqrt{\alpha_t}\x_{t-1} + \sqrt{1-\alpha_t}\boldsymbol{\epsilon}_t\nonumber
\end{align}
where $\alpha_t = 1 - \beta_t$ and $\boldsymbol{\epsilon}_t\sim\mathcal{N}(\mathbf{0,I})$, producing a sequence of noisy samples. The step sizes are controlled by a variance schedule ${\{\beta_t\in(0,1)\}}_{t=1}^T$.
As $t$ becomes larger, the image gradually loses its distinguishable features. When $t\rightarrow \infty$ , $\x_{t}$ becomes a sample of an isotropic Gaussian distribution.

If we can reverse the above process and sample from $q(\x_{t-1} | \x_{t})$, we will be able to recreate the true sample from a Gaussian noise $\x_T\sim\mathcal{N}(\mathbf{0},\mathbf{I})$. 
If $\beta_t$ is small enough, $q(\x_{t-1} | \x_{t})$ will also be a Gaussian. So we can approximate it with a parameterized model, as shown in \cref{eqn:reverse_q}
\begin{equation}
    \label{eqn:reverse_q}
    p_{\boldsymbol{\theta}}(\x_{t-1} | \x_{t})=\mathcal{N}(\x_{t-1};\boldsymbol{\mu_{\theta}}(\x_t,t),\boldsymbol{\Sigma_{\theta}}(\x_t,t)) \;.
\end{equation}
Since $q(\x_{t-1} | \x_{t},\x_0)$ is tractable,
\begin{equation}
    q(\x_{t-1} | \x_{t},\x_0)=\mathcal{N}(\x_{t-1};\tilde{\boldsymbol{\mu}}(\x_t,\x_0),\tilde{\beta_t}\mathbf{I})
\end{equation}
where $\bar{\alpha}_t = \prod_{s=1}^t \alpha_s$, and
\begin{align}
    \tilde{\boldsymbol{\mu}}(\x_t,\x_0) & = \frac{1}{\sqrt{\alpha_t}}(\x_t - \frac{1-\alpha_t}{\sqrt{1-\bar{\alpha}_t}}\boldsymbol{\epsilon}_t)\\
    \tilde{\beta_t}&=\frac{1-\bar{\alpha}_{t-1}}{{1-\bar{\alpha}_t}}\cdot\beta_t \;.
\end{align}

So we can train a neural network to approximate $\boldsymbol{\epsilon}_t$ and the predicted value is denoted as $\boldsymbol{\epsilon_{\theta}}(\x_t)$. It has been verified that instead of directly setting $\boldsymbol{\Sigma_{\theta}}(\x_t,t)$ as $\tilde{\beta_t}$, setting it as a learnable interpolation between $\tilde{\beta_t}$, $\beta_t$ in log domain will yield better log-likelihood \cite{nichol2021improved}:
\begin{equation}
    \boldsymbol{\Sigma_{\theta}}(\x_t,t) = \exp(\boldsymbol{\nu}_{\theta}(\x_t)\log{\beta_t}
    +(1-\boldsymbol{\nu}_{\theta}(\x_t))\log{\tilde{\beta_t}}) \;.
\end{equation}
In this paper, we will train a U-Net to predict $\boldsymbol{\epsilon_{\theta}}(\x_t)$ and $\boldsymbol{\nu}_{\theta}(\x_t)$ with the same hybrid loss as in \cite{nichol2021improved}.

\subsection{Glyph Conditional U-Net Architecture}
\label{sec:cdn-unet}
As shown in \cref{fig:unet}, the U-Net architecture we used is borrowed from \cite{dhariwal2021diffusion}. 
With $128\times128$ image input, there are 5 resolution stages in encoder and decoder respectively, and each stage consists of 2 BigGAN residual blocks (ResBlock) \cite{brock2018large}. 
In addition, BigGAN ResBlocks are also used for downsampling and upsampling activations.
We also follow \cite{dhariwal2021diffusion} to use multi-head attention at $32\times32$, $16\times16$ and $8\times8$ resolutions. 
Timestep $t$ will first be mapped to sinusoidal embedding and then processed by a 2-layer feed-forward network (FFN). This processed embedding will then be fed to each convolution layer in U-Net through a feature-wise linear modulation (FiLM) operator \cite{perez2018film}.

To control the style and content of generated character images, writer information ~\cite{graves2013generating} and character category information are also fed to the model. 
Given a writer $\mathbf{w}$, 
which is actually the class index of all writer IDs,
it will be mapped to a learnable embedding, followed by L2-normalization (denoted as $\mathbf{z}$), which is injected to U-Net together with the timestep embedding \cite{nichol2021improved} as shown in \cref{fig:unet}.

If we inject character category information in the same way as writer, the model will not be able to generate samples for unseen categories because their embeddings are not optimized at all. 
In this paper, we propose to leverage printed images rendered by font ``kai'' to provide character category information. We denote this glyph image as $\mathbf{g}$. 
There are several ways to inject $\mathbf{g}$ to the model. 
For example, it can be encoded as a feature vector by a CNN/ViT and fed to U-Net in FiLM way, or encoded as feature sequences and fed to attention layers of U-Net serving as external keys and values \cite{nichol2021glide}. 
In this paper, we simply inject $\mathbf{g}$ as model's input by concatenating it with $\x_t$ and leave other ways as future work. We call our approach as \textbf{G}lyph \textbf{C}onditional DDPM (\textbf{GC-DDPM}).

By conditioning model output on glyph image, we expect the model can learn the implicit mapping rules between printed stroke combinations and their handwritten counterparts. Then we can input font-rendered glyph images of unseen characters to the well-trained GC-DDPM and get their handwritten samples of high quality and diversity.

\subsection{Multi-conditional Classifier-free Diffusion Guidance}
\label{sec:mcs}
Classifier-free guidance \cite{ho2022classifier} has been proven effective for  improving generation quality on different tasks. In this paper, we are also curious about its effects on HCCR system trained with synthetic samples.

There are 2 conditions, glyph $\mathbf{g}$ and writer $\w$, in our model. We assume that given $\x_t$, $\mathbf{g}$ and $\w$ are independent. So we have
\begin{align}
    p_{\boldsymbol{\theta}}(\x_{t-1}|\x_{t},\g,\w ) 
    & \propto  p_{\boldsymbol{\theta}}(\x_{t-1} | \x_{t}) p_{\boldsymbol{\theta}}(\g | \x_{t})p_{\boldsymbol{\theta}}(\w | \x_{t}) \;.
\end{align}

Following the previous practice in \cite{ho2022classifier}, we assume that there is an implicit classifier (ic), 
\begin{align}
    p_{ic}(\g,\w|\x_t)\propto \left[\frac{p(\x_t|\g)}{p(\x_t)}\right]^\gamma\cdot
    \left[\frac{p(\x_t|\w)}{p(\x_t)}\right]^\eta \;.
\end{align}
Then we have
\begin{align}
    \nabla_{\x_t} \log p_{ic}(\g,\w|\x_t) \propto \gamma \boldsymbol{\epsilon}(\x_t,\g) + \eta\boldsymbol{\epsilon}(\x_t,\w) - (\gamma+\eta)\boldsymbol{\epsilon}(\x_t) \;.
\end{align}
So we can perform sampling with the score formulation
\begin{equation}
    \begin{aligned}
    \tilde{\boldsymbol{\epsilon}}_{\boldsymbol{\theta}}(\x_t, \g,\w)
    &=\boldsymbol{\epsilon}_{\boldsymbol{\theta}}(\x_t, \g,\w)+
    \gamma \boldsymbol{\epsilon}_{\boldsymbol{\theta}}(\x_t, \g,\emptyset )\\
    &+
    \eta \boldsymbol{\epsilon}_{\boldsymbol{\theta}}(\x_t, \emptyset,\w )
    -
    (\gamma+\eta)\boldsymbol{\epsilon}_{\boldsymbol{\theta}}(\x_t, \emptyset,\emptyset )  \;.
\end{aligned}
\label{eq:mc cf}
\end{equation}
We call $\gamma$, $\eta$ as content and writer guidance scales respectively. When $\g=\emptyset$, an empty glyph image will be fed to U-Net and when $\w=\emptyset$, a special embedding will be used. 
During training, we set $\g$ and $\w$ to $\emptyset$ with probability 10\% independently to get partial/unconditional models.

\subsection{Writer Interpolation}
\label{sec:style interploation}
Besides generating unseen characters, our model is also able to generate unseen styles by injecting interpolation between different writer embeddings as new writer embedding. 
Given two normalized writer embeddings $\mathbf{z_i}$ and $\mathbf{z_j}$, we use spherical interpolation \cite{ramesh2022hierarchical} to get a new embedding $\mathbf{z}$ with L2-norm being 1, as in  \cref{eq:spherical}:
\begin{align}
\label{eq:spherical}
\mathbf{z} & = \mathbf{z}_{i} \cos \frac{\lambda \pi}{2}+\mathbf{z}_{j} \sin \frac{\lambda \pi}{2}, \quad \lambda \in[0,1] \;.
\end{align}

%% file: 4experiments.tex
\section{Experiments}
\label{sec:experiments}

We conduct our experiments on CASIA-HWDB~\cite{liu2011casia} dataset. The detailed experimental setup is comprehensively explained in ~\cref{sec:exp-setup}. Experiments on Writer Independent (WI) and Writer Dependent (WD) GC-DDPMs are conducted in ~\cref{sec:exp-wi-ddpm} and ~\cref{sec:exp-wd-ddpm}, respectively. We further use synthesized samples to augment the training set of HCCR in ~\cref{sec:exp-augment}. Finally, we compare our approach with prior arts in ~\cref{sec:exp-comp}.

\subsection{Experimental Setup} \label{sec:exp-setup}

\textbf{Dataset}: The CASIA-HWDB dataset is a large-scale offline Chinese handwritten character database including HWDB1.0, 1.1 and 1.2. We use the HWDB1.0 and 1.1 in experiments, where the former contains 3,866 Chinese character categories written by 420 writers, and the latter contains 3,755 categories written by another 300 writers. We follow the official partition of training and testing sets as in ~\cite{liu2011casia}, where the training set is written by 576 writers.

\textbf{Vocabulary partition}: We use the 3,755 categories that cover the standard GB2312-80 level-1 Chinese set in experiments. We denote the set of 3,755 categories as $\mathcal{S}_{3,755}$. Following the setup in ~\cite{ao2019cross,wang2018denseran}, we select the first 2,000 categories in GB2312-80 set as seen categories (denoted as $\mathcal{S}_{2,000}$), and the remaining 1,755 categories as unseen categories (denoted as $\mathcal{S}_{1,755}$). The diffusion models are trained on training samples of $\mathcal{S}_{2,000}$ and used to generate handwritten Chinese character samples of  $\mathcal{S}_{1,755}$ to evaluate the performance of zero-shot training data generation for HCCR.

\textbf{DDPM settings}: Our DDPM implementation is based on ~\cite{dhariwal2021diffusion}. We use the ``kai'' as our font library to render printed character images. 
We conduct experiments on both WI and WD GC-DDPMs. 
In WI GC-DDPM training, we disable writer embeddings and randomly set content condition $\g$ as $\emptyset$ with probability 10\%. 
And in WD GC-DDPM, writer condition $\w$ is also randomly set to $\emptyset$ with  probability 10\%. 
Flip and mirror augmentations are used during training. We set batch size as 256, image size as 128$\times$128, and we use AdamW optimizer~\cite{loshchilov2017decoupled} with learning rate 1.0e-4. 
Diffusion step number is set to 1,000 with a linear noise schedule.
GC-DDPMs are trained for about 200K steps using a machine with 8 Nvidia V100 GPUs, which takes about 5 days. 
During sampling, we use the denoising diffusion implicit model (DDIM)~\cite{song2020denoising} sampling method with 50 steps.
It takes 62 hours to sample 3,755 characters written by 576 writers, which are about 2.2M samples, with the same 8 Nvidia V100 GPUs. 

\textbf{Evaluation metrics}: We evaluate the quality of synthetic samples in three aspects. First, Inception score (IS)~\cite{NIPS2016_8a3363ab} and Frechet Inception Distance (FID)~\cite{heusel2017gans} are used to evaluate the diversity and distribution similarity of synthetic samples compared with real ones. Second, since samples are synthesized by conditioning on glyph image, the synthetic samples should be consistent with the category of conditioned glyph. Therefore, we introduce a new metric called correctness score (CS). For each synthetic sample, the category of conditioned glyph is used as ground truth, and CS is calculated as the recognition accuracy of synthetic samples using an HCCR model trained with real data,
which achieves 97.3\% recognition accuracy in real data testing set. 
Finally, as the purpose of diffusion model here is to generate training data for unseen categories, we also train HCCR models with synthetic samples and evaluate recognition accuracy on the real testing set of unseen categories. Our HCCR model adopts ResNet-18~\cite{he2016deep} architecture and is trained with standard SGD optimizer. No data augmentation is applied during HCCR model training. 
It is noted that starting from different random noise, it is almost impossible to generate exact same handwritten samples even for same conditional character glyphs. So it is not appropriate to adopt pixel-level metrics to evaluate generative effect as ~\cite{huang2020rd,park2021few,liu2022xmp,zhang2018separating,gao2019artistic,zhu2020few,xie2021dg,liu2022fonttransformer} do.

%

\subsection{WI GC-DDPM Results} \label{sec:exp-wi-ddpm}

\begin{figure}[!t]
	\centering
    \includegraphics[width=0.7\linewidth]{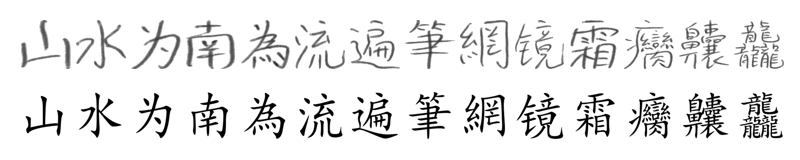}

	\caption{Synthetic handwritten Chinese character samples and corresponding glyphs, with stroke numbers increasing from left to right.}
	\label{fig:show}
    \vspace{-3mm}
\end{figure}

\begin{table}[t]
    \begin{center}
    \caption{Comparisons of generation quality using different content guidance scale $\gamma$'s in terms of IS, FID, and CS. }
        \vspace{2mm}
        \begin{tabular}{|c|c|c|c|}
            \hline
            $\gamma$ & IS & FID & CS (\%) \\  
            \hline            
            0.0 & 2.62 & 8.07 & 94.7  \\ 
            \hline            
            1.0 & 2.51 & 10.97 & 99.8\\ 
            \hline            
            2.0 & 2.46 & 18.03 & 99.9 \\ 
            \hline            
            3.0 & 2.44 & 24.34 & 99.9 \\ 
            \hline            
            4.0 & 2.39 & 28.69 & 99.9 \\ 
            \hline
        \end{tabular}
        \vspace{-2mm}
        \label{table:ind_diversity}
    \end{center}
\end{table}

\begin{table}[t]
    \begin{center}
    \caption{Comparisons of generation quality using different content guidance scale $\gamma$'s in terms of recognition accuracy on testing set of classes in $\mathcal{S}_{1,755}$ using generated samples as training set.}
        \vspace{2mm}
        \begin{tabular}{|c|c|c|c|c|c|}
            \hline
            $\gamma$ & 0.0 & 1.0 & 2.0 & 3.0 & 4.0 \\
            \hline
            $\text{Acc}_{1,755}$ (\%) & 93.0 & 88.6 & 91.7 & 63.7 & 33.2\\
            \hline
        \end{tabular}
        \label{table:ind_acc}
        \vspace{-2mm}
    \end{center}
\end{table}

\begin{figure}[!t]
	\centering
    \subfloat[Failure samples that do not look like any Chinese characters.]{
    \label{fig:error a}
        \includegraphics[width=0.55\linewidth]{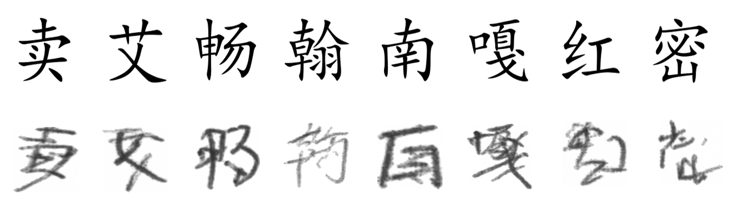}
    }
   \vspace{0mm}
	\subfloat[ (top) Glyph condition images; 
	(middle) Synthetic samples;
	(bottom) Most similar characters.
	]{
    \label{fig:error b}
        \includegraphics[width=0.55\linewidth]{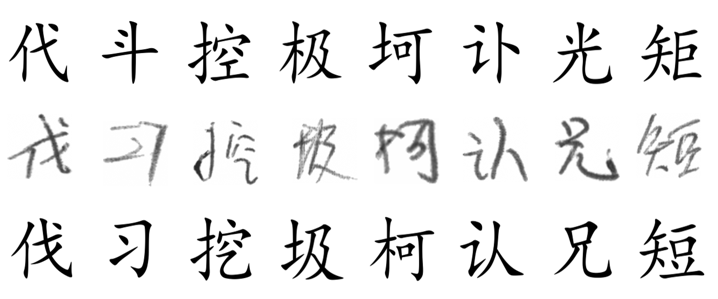}
    }
	
	\caption{Synthetic samples that are wrongly recognized by real data trained HCCR model when $\gamma=0$.}
	\label{fig:failure}
    \vspace{-5mm}
\end{figure}

\begin{figure}[!t]
	\centering
 
    \subfloat[$\mathcal{S}_{2,000}$ example]{
    \label{fig:ind_guidance a}
        \includegraphics[width=0.5\linewidth]{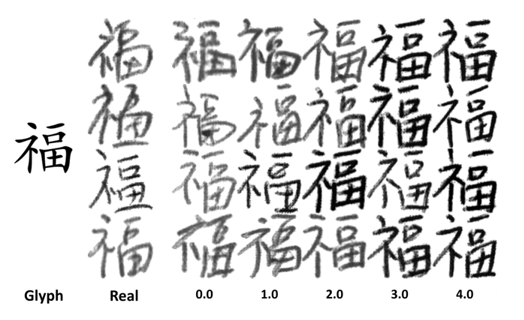}
    }
    \subfloat[$\mathcal{S}_{1,755}$ example]{
    \label{fig:ind_guidance b}
	\includegraphics[width=0.5\linewidth]{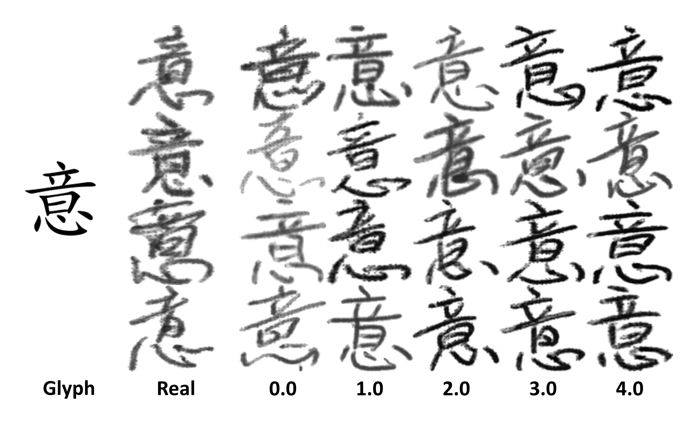}
    }
    \vspace{0mm}
    \subfloat[Out of $\mathcal{S}_{3,755}$ example]{
    \label{fig:ind_guidance c}
	\includegraphics[width=0.5\linewidth]{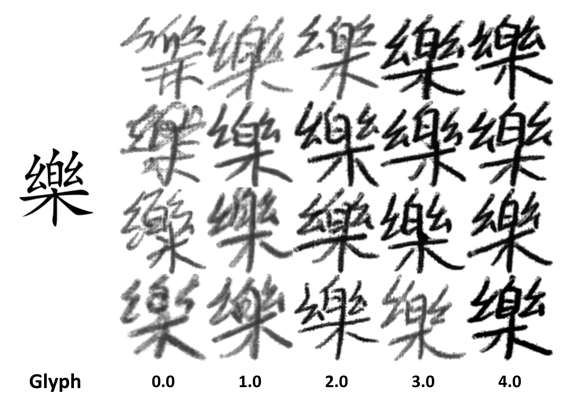}
    }
    \subfloat[Complicated strokes example]{
    \label{fig:ind_guidance d}
	\includegraphics[width=0.5\linewidth]{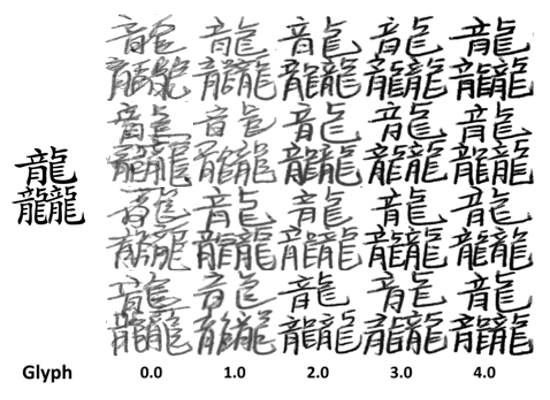}
    }
	\caption{Multiple synthetic handwritten Chinese character samples with different content guidance scale, where (a), (b) and (c) are characters from classes of $\mathcal{S}_{2,000}$, $\mathcal{S}_{1,755}$, and out of $\mathcal{S}_{3,755}$ Chinese character sets. Samples in each line use the same random seed and initial noise. Samples across lines use different random seeds to visualize diversity.}
	\label{fig:ind_guidance}
    \vspace{-5mm}
\end{figure}

We first conduct experiments on WI GC-DDPM. It is shown in~\cite{ho2022classifier} that the classifier guidance scale is able to attain a trade-off between quality and diversity. In order to evaluate the behavior of different content guidance scale $\gamma$'s, we choose different $\gamma$'s and generate samples to compute FID, ID and CS. 
Here we synthesize 50K samples of  $\mathcal{S}_{2,000}$, and the HCCR model used to measure CS is trained using real samples of $\mathcal{S}_{3,755}$. $\gamma \in \{ 0.0,~1.0,~2.0,~3.0~,4.0\}$ are used and the comparison results are summarized in \cref{table:ind_diversity}. We can find that, as $\gamma$ increases, the IS decreases, the FID increases and the CS achieves close to 100\% accuracy. This indicates that with a larger $\gamma$, the diversity of synthetic samples is decreasing. This behavior is also observed in \cref{fig:ind_guidance a} where we visualize multiple sampled results of the character class in  $\mathcal{S}_{2,000}$ using different $\gamma$'s. The generated samples are less diverse, less cursive and easier to recognize when conditioned on stronger content guidance. According to FID and examples in \cref{fig:ind_guidance}, the distribution of synthetic samples with $\gamma=0$ is closer to that of real samples. When $\gamma=0$, CS achieves $94.7\%$. 
In \cref{fig:failure}, we show synthetic cases that the trained HCCR model fails to recognize. Failure cases include (a) samples that are unreadable, and (b) samples that are closer to another easily confused Chinese character. 
They are caused by alignment failures between  printed and synthetic strokes, and can be eliminated by improving glyph conditioning method. 
We leave it as future work.

Then, we evaluate the quality of WI GC-DDPM for zero-shot generation of HCCR training data. We use the trained WI GC-DDPM to synthesize 576 samples for each category in $\mathcal{S}_{1,755}$. Then, the synthetic samples are used along with real samples of categories in $\mathcal{S}_{2,000}$ to train an HCCR model that supports 3,755 categories. We calculate its recognition accuracy on the testing set of category $\mathcal{S}_{1,755}$, which is denoted as $\text{Acc}_{1,755}$. Different $\gamma$'s are tried, and the results are shown in \cref{table:ind_acc}. In \cref{fig:ind_guidance b}, we visualize synthetic samples of one category in $\mathcal{S}_{1,755}$. The best $\text{Acc}_{1,755}$ is achieved when $\gamma=0$. Although synthetic samples with higher $\gamma$ are less cursive, they achieve much lower $\text{Acc}_{1,755}$. This is because the lack of diversity makes it difficult to cover the wide distribution of handwritten Chinese character image space.

Clearly, by learning the mapping of radicals and spatial relationship between Chinese printed and handwritten strokes, the diffusion model is capable of zero-shot generation of unseen Chinese character categories. Moreover, a high accuracy of $93.0\%$ is achieved on $\mathcal{S}_{1,755}$ by only leveraging the synthetic samples. In \cref{fig:ind_guidance c,fig:ind_guidance d}, we further show the synthetic samples of a Chinese character category that does not belong to $\mathcal{S}_{3,755}$. 
The excellent generation effect implies that our method has the potential to be extended to a larger vocabulary.

\subsection{WD GC-DDPM Results} \label{sec:exp-wd-ddpm}

\begin{figure}[t]
	\centering	\includegraphics[width=0.5\linewidth]{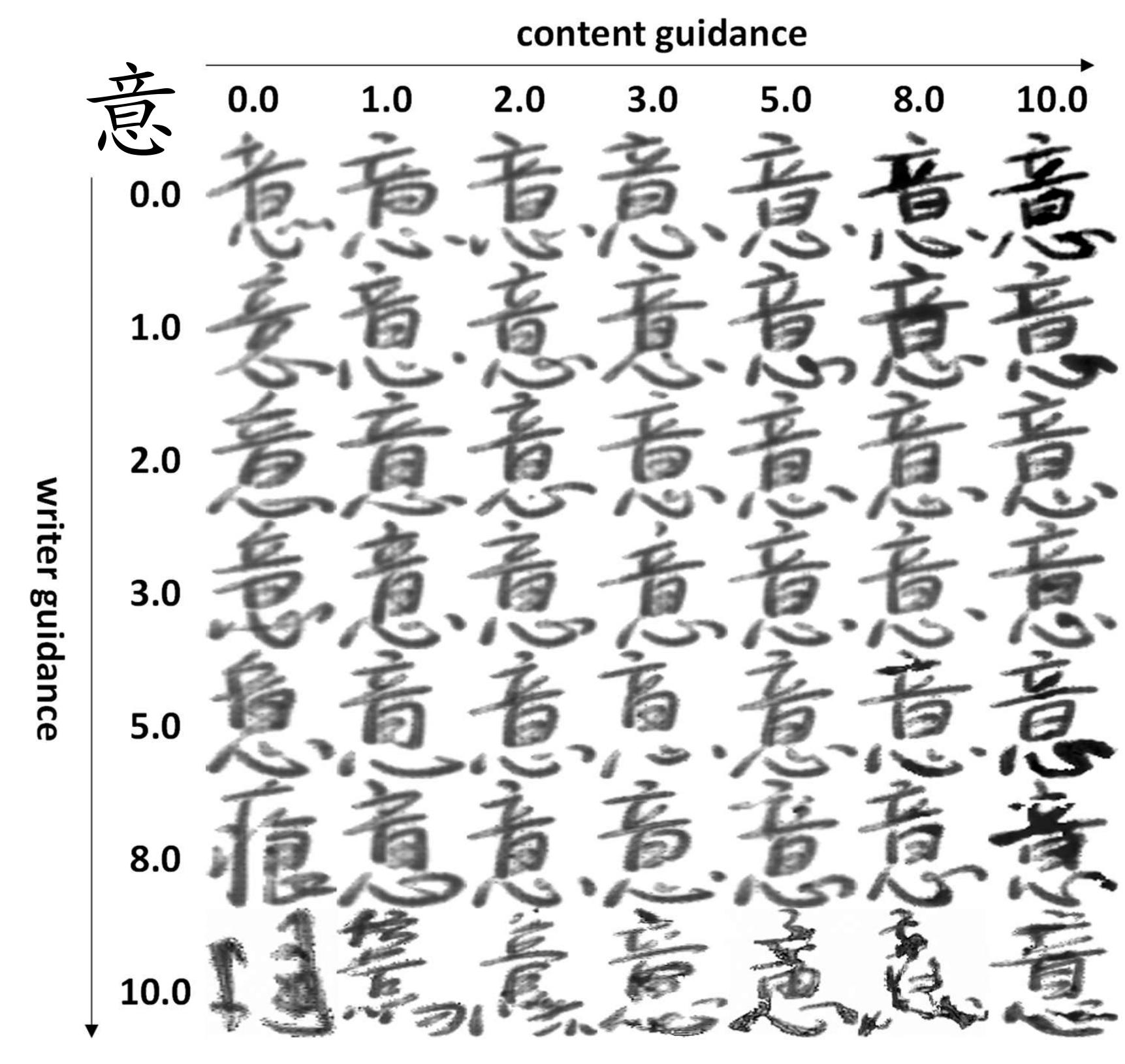}
	\caption{Generated handwritten Chinese character samples with different content and writer guidance scales, where the character is from the class of $\mathcal{S}_{1,755}$. Samples are generated with the same random seed and initial noise.}
	\label{fig:dep_guidance}
\end{figure}

\begin{figure*}[t]
	\centering
    \centering
    \subfloat[Real text line from \cite{liu2011casia}.]{
    \label{fig:textline raw}
        \includegraphics[width=0.65\linewidth]{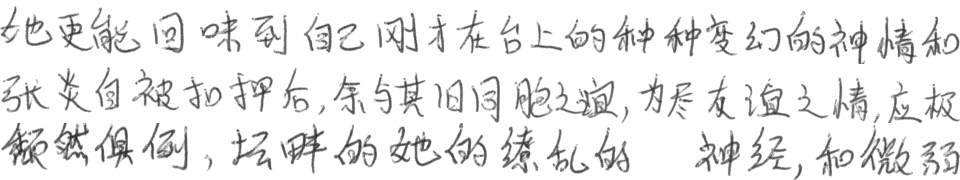}
    }
    \vspace{0mm}
	\subfloat[Synthetic samples arranged as a text line. ]{
    \label{fig:textline syn}
        \includegraphics[width=0.65\linewidth]{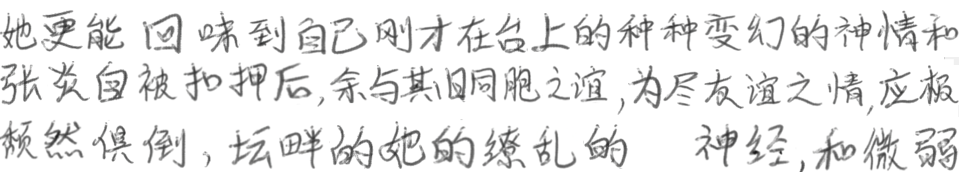}
    }
	
     \caption{Comparisons of real text line images in HWDB2.1 and generated samples arranged in a text line, where we replace the characters from real data with the generated characters. Samples in different lines of (a) and (b) are selected and generated conditioning on the same writer 1001.}
     \label{fig:textline}
\end{figure*}

\begin{figure*}[t]
	\centering
 
	\includegraphics[width=0.7\linewidth]{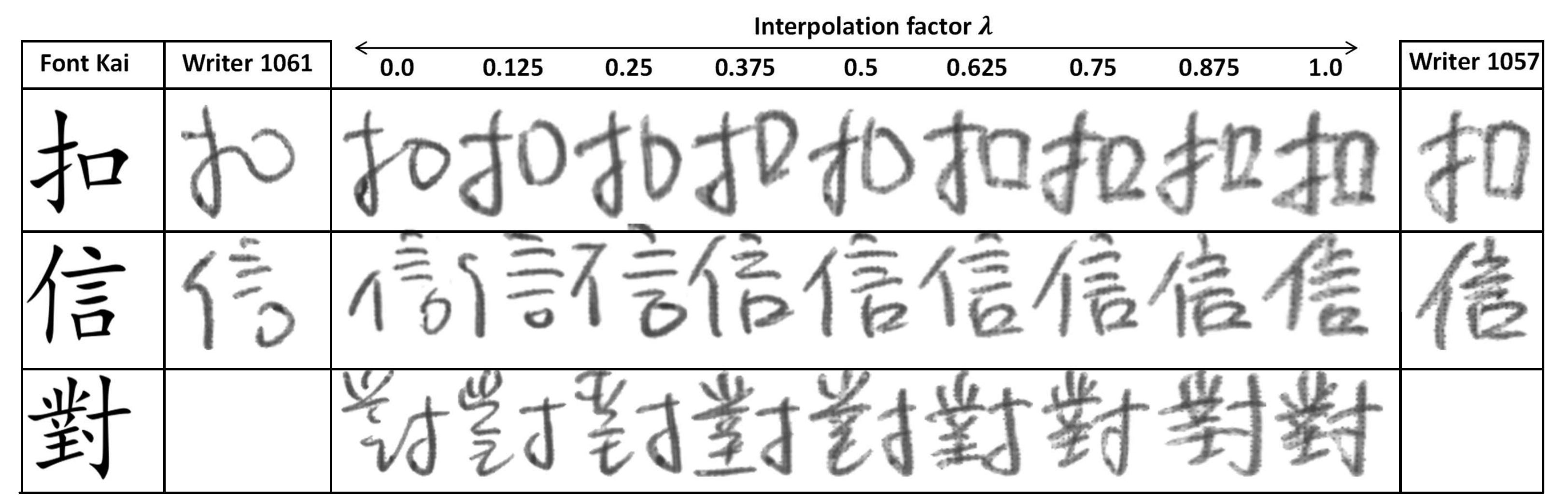}
	\caption{Interpolation of handwritten Chinese character samples, where the top, middle, bottom lines are characters from classes of $\mathcal{S}_{2,000}$, $\mathcal{S}_{1,755}$, and out of $\mathcal{S}_{3,755}$ Chinese character sets. We choose writer 1061 (left) and writer 1057 (right) for interpolation and interpolation factors are shown at the top of images. Standard glyph images of font ``kai" are shown on the left. Samples in each line use the same random seed and initial noise.}
	\label{fig:interpolation}
 \vspace{-4mm}
\end{figure*}

\begin{table}[t]
    \begin{center}
    \caption{Comparisons of generation quality between WI and WD DDPMs in terms of IS, FID, CS (\%) and the recognition accuracy (\%) on the testing set of class $\mathcal{S}_{1,755}$ using generated samples as training set. }
         \vspace{2mm}
        \begin{tabular}{|c|c|c|c|c|}
            \hline
            Model & IS & FID & CS & $\text{Acc}_{\text{1,755}}$ \\
            \hline            
            WI & 2.62 & 8.07 & 94.7& 93.0  \\ 
            \hline           
            WD & 2.49 & 6.34 & 94.8 & 93.7 \\ 
            \hline    
            \hline
            WD w/ interpolation& 2.53 & 6.26 & 95.0 & 94.7 \\ 
            \hline
        \end{tabular}
        \label{table:dep}
        \vspace{-2mm}
    \end{center}
\end{table}
Although WI GC-DDPM can generate desired handwritten characters, we cannot control their writing styles. In this part, we conduct experiments on WD GC-DDPM, which introduces writer information as an additional condition.

\cref{fig:dep_guidance} shows the visualization results of sampling with 
different content guidance scale $\gamma$'s and writer guidance scale $\eta$'s.
It shows that with larger $\gamma$, the synthetic samples become less cursive and more similar to the corresponding printed image. This behavior is consistent with that of the WI GC-DDPM in \cref{fig:ind_guidance}. We also find that with large $\eta$, the generated sample becomes inconsistent with the conditioned printed image.
Since writer information is injected to GC-DDPM in FiLM way, a large guidance scale will  cause the mean and variance shift of $\tilde{\boldsymbol{\mu}}_{\boldsymbol{\theta}}(\x_t, \g,\w)$ and $\tilde{\boldsymbol{\Sigma}}_{\boldsymbol{\theta}}(\x_t, \g,\w)$ which
hinders the subsequent denoising, leading to over-saturated images with over-smoothed textures \cite{wang2022pretraining}.

In \cref{fig:textline syn}, we show several synthetic text line images conditioned on a fixed writer embedding with our WD GC-DDPM. Writing styles of these samples are consistent and quite similar to real samples written by the same writer as shown in \cref{fig:textline raw}. These results verify the writing style controllability of our model.

Then, we compare the quality of synthetic samples when used as training data for HCCR. For a fair comparison, we also generate 576 samples for each category in $\mathcal{S}_{1,755}$, one image for each writer. Recognition performances are shown in \cref{table:dep}. To 
improve sampling efficiency and 
ensure training data diversity, the writer guidance scale of 0 is applied. Compared with using samples synthesized with WI GC-DDPM as HCCR training set, the accuracy on the testing set of $\mathcal{S}_{1,755}$ is improved from $93.0\%$ to $93.7\%$. When GC-DDPM is trained without conditioning on writer embedding, it may generate similar samples from different initial noise. Whereas in WD GC-DDPM, by conditioning on different writer embeddings, the model will generate samples with different writing styles. Therefore, the  diversity of synthetic samples will be improved. To verify this, we compare the quality of synthetic samples in terms of IS and FID. As shown in \cref{table:dep}, the FID improves from 8.07 to 6.34. The results demonstrate the superiority of WD GC-DDPM in zero-shot training data generation of unseen Chinese character categories.

Another capability of WD GC-DDPM is that it can interpolate between different writer embeddings and generate samples of new styles. We choose 2 writers and try different interpolation factor $\lambda$'s and visualize the synthetic samples in \cref{fig:interpolation}. We find that as $\lambda$ increases from 0 to 1, the style of synthetic samples gradually shifts from one writing style to another. We also observe that with the same $\lambda$, the synthetic samples of different Chinese characters share similar writing style as expected. Finally, we use writer style interpolation to generate the training data of $\mathcal{S}_{1,755}$ for HCCR, and again 576 samples are generated for each category. For each image, we randomly select 2 writers for interpolation. We simply use an interpolation factor of 0.5. Results are summarized in \cref{table:dep}. We observe a slight improvement in FID score and a $1\%$ absolute recognition accuracy improvement on $\mathcal{S}_{1,755}$, which further verifies the superiority of our WD GC-DDPM.

\subsection{Data-augmented HCCR Results} \label{sec:exp-augment}

\begin{table}[t]
    \begin{center}
    \caption{Comparisons of recognition accuracy (\%) on test sets of  $\mathcal{S}_{2,000}$ and $\mathcal{S}_{1,755}$ using real and/or synthetic samples as HCCR training set. }
        \begin{tabular}{|c|c|c|c|}
            \hline
            \multicolumn{2}{|c|}{Training set} & \multicolumn{2}{c|}{Accuracy on testing set} \\ \hline
            Real & Synthetic & $\text{Acc}_{\text{2,000}}$ & $\text{Acc}_{\text{1,755}}$  \\ \hline
            \checkmark & / & 97.3 & 97.2 \\ \hline \hline
            / & WI & 96.3 & 96.0 \\ \hline
            / & WD & 96.4 & 96.1 \\ \hline
            / & WD w/ interpolation & 96.5 & 96.1 \\ \hline \hline
            \checkmark & WI & 97.3 & 97.3 \\ \hline
            \checkmark & WD & 97.4 & 97.3 \\ \hline
            \checkmark & WD w/ interpolation & 97.4 & 97.3 \\ \hline
        \end{tabular}
        \label{table:aug}
        \vspace{-4mm}
    \end{center}
\end{table}
We also use GC-DDPMs trained on $\mathcal{S}_{2,000}$, to synthesize samples for all categories in $\mathcal{S}_{3,755}$, and combine them with real samples to build HCCR systems.
3 settings are tried: WI, WD and WD w/ interpolation. And 576 samples for each category are synthesized in each setting.
\cref{table:aug} summarizes the results.
Best accuracies are achieved with samples synthesized by WD w/ interpolation, which is consistent with \cref{table:dep}. 
The HCCR models trained with only synthetic samples perform slightly worse than the one trained with only real samples.
Combining synthetic and real training samples only performs 0.0\%\textasciitilde0.1\% better than real samples.
These results demonstrate the distribution modeling capacity of GC-DDPMs.

\begin{table}[!t]
    \begin{center}
    \caption{ Comparisons of unseen character categories' recognition accuracy (\%) between our method and prior zero-shot HCCR systems. 
    Works with ${}^*$ also use samples from HWDB1.2 for training, while ${}^\dag$ means online trajectory information is also used.
    }
        \begin{tabular}{|c|c|}
            \hline
            Method&Accuracy\\
            \hline
            CM${}^\dag$~\cite{ao2019cross}&86.7\\
            \hline
            DenseRan~\cite{wang2018denseran}&19.5\\
            \hline
            FewRan${}^*$~\cite{wang2019radical}&70.6\\
            \hline
            HCCR${}^*$~\cite{cao2020zero}&73.4\\
            \hline
            OSOCR${}^*$~\cite{liu2022towards}&84.3\\
            \hline
            OSCCD${}^*$~\cite{liu2022open}&95.6\\
            \hline 
            \hline
            WI GC-DDPM &96.4\\
            \hline
            WD GC-DDPM &96.8\\
            \hline
            WD GC-DDPM w/ interpolation &\textbf{96.9}\\
            \hline
        \end{tabular}
        \label{table:recog}
        \vspace{-6mm}
    \end{center}
\end{table}

\begin{table}[!t]
    \begin{center}
    \caption{ Comparisons of unseen character categories' recognition accuracy (\%) on CASIA1.2 testing set.}
        \begin{tabular}{|c|c|}
            \hline
            Methods&Accuracy\\
            
            \hline            
            RCN ~\cite{xue2021radical}&46.1\\
            
            \hline            
            \hline            
            WI GC-DDPM &98.6\\
            \hline            
            WD GC-DDPM &98.6\\
            \hline 
            \hline            
            ResNet-18 trained with real data &97.9\\
            \hline
        \end{tabular}
        \label{table:compare du}
        \vspace{-5mm}
    \end{center}
\end{table}

\subsection{Comparison with Prior Arts} \label{sec:exp-comp}
\label{sec:comparison}

Finally, we compare our method with prior arts. We first compare our method with prior zero-shot HCCR systems. To be consistent with prior works in ~\cite{cao2020zero,liu2022towards,liu2022open}, we randomly choose 1,000 classes in $\mathcal{S}_{1,755}$ as unseen classes and use ICDAR2013~\cite{yin2013icdar} benchmark dataset for testing. Results are shown in \cref{table:recog}. Here we only list the results from prior arts using 2,000 seen character classes. It is noted that the 2,000/1,000 seen/unseen character class split for training and testing is not exactly the same. So the results are not directly comparable. The results in \cref{table:recog} show that our methods achieve the same level recognition accuracy compared with previous state-of-the-art zero-shot HCCR systems. Moreover, our approach directly uses a standard CNN to predict supported categories, which is much simpler compared with the systems in ~\cite{liu2022towards,liu2022open}.

We also compare our approach with ~\cite{xue2021radical}, which also leverages a generation model to synthesize training samples for unseen classes. We follow the same experimental setups in ~\cite{xue2021radical} and use HWDB1.0 and 1.1 as training set, which contains 3,755 categories, to train GC-DDPMs. Unseen 3,319 categories in HWDB1.2 testing set are used as testing set. Results are shown in \cref{table:compare du}. ~\cite{xue2021radical} achieves a $46.1\%$ accuracy by adding more than 9.6M generated samples. Our approach achieves a $98.6\%$ accuracy by only adding about 1.9M synthetic samples (576 samples for each unseen category). We also train a classifier using all real samples in HWDB1.2 training set (240 samples for each category). The classifier achieves a $97.9\%$ accuracy, which is slightly worse than ours due to less diverse training samples.

These results verify the zero-shot generation capability of our methods again. It is easy to extend to larger vocabularies, which makes it possible to build a high-quality HCCR system for 87,887 categories.

%% file: 5future.tex
\section{Limitations and Future Work}
\label{sec:future}
\begin{figure}[!t]
	\centering
    \subfloat[Japanese]{
    \label{fig:japan}
        \includegraphics[width=0.4\linewidth]{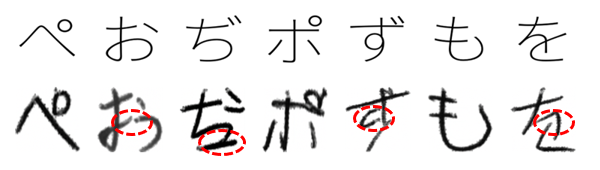}
    }
	\subfloat[Korean]{
    \label{fig:korea}
        \includegraphics[width=0.4\linewidth]{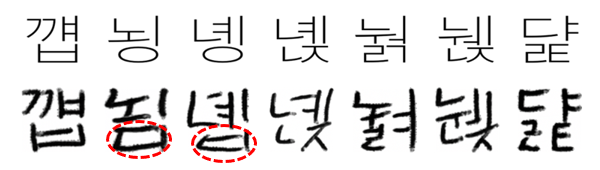}
    }
	
    \caption{Synthetic samples of Japanese and Korean characters and standard glyph images in font ``SourceHans".}
	\label{fig:jk}
    \vspace{-5mm}
\end{figure}

Although GC-DDPM-synthesized images are quite helpful for building a high-quality HCCR system, there are still some failure cases.
The blur and dislocation phenomena in these samples reveal that there exist better ways to inject glyph information. 
It is also possible to encode radical/stroke sequences with spatial relationships as the condition of DDPM.
We will investigate these methods and report the results elsewhere.

Another limitation of our approach is the long training time of DDPMs. We will try to reduce the number of character categories and sample numbers per category to find a better trade-off between synthesis quality and training cost.

Japanese and Korean characters share most strokes with Chinese, so we also try to synthesize handwritten Japanese and Korean samples with our Chinese-trained DDPM.
As \cref{fig:jk} shows, except for some circle and curve strokes, the results are quite reasonable. As future work, we will combine handwritten samples of CJK languages to build a new DDPM, which is expected to synthesize samples for each language with higher diversity and quality.

%% file: 6conclusion.tex
\section{Conclusion}
\label{sec:conclusion}
We propose WI and WD GC-DDPM solutions to achieve zero-shot training data generation for HCCR.
Experimental results have verified their effectiveness in terms of generation quality, diversity and HCCR accuracies of unseen categories. WD performs slightly better than WI due to its better distribution modeling capability and writing style controllability.
These solutions can be easily extended to larger vocabularies and other languages, and provide a feasible way to build an HCCR system supporting 87,887 categories with high recognition accuracy.